%% file: main.tex
\pdfoutput=1

\documentclass[11pt]{article}

\usepackage{EMNLP2023}
\usepackage{amsmath}
\usepackage{url}
\usepackage{amssymb}
\usepackage{amsfonts}
\usepackage{graphicx}
\usepackage{tabularx}
\usepackage{multirow}
\usepackage{arydshln}
\usepackage{mathtools,nccmath}
\usepackage{enumitem}
\usepackage{todonotes}
\usepackage{times}
\usepackage{latexsym}
\usepackage{subcaption}
\usepackage{caption}
\DeclareMathOperator*{\argmax}{argmax}

\usepackage[T1]{fontenc}

\usepackage[utf8]{inputenc}

\usepackage{microtype}

\usepackage{inconsolata}

\title{MISCA: A Joint Model for Multiple Intent Detection and Slot Filling\\ with Intent-Slot Co-Attention}

\author{Thinh Pham \and Chi Tran \and Dat Quoc Nguyen  \\
  VinAI Research, Vietnam \\
  \texttt{\{v.thinhphp1, v.chitb, v.datnq9\}@vinai.io}}

\begin{document}
\maketitle
\begin{abstract}

The research study of detecting multiple intents and filling slots is becoming more popular because of its relevance to complicated real-world situations. Recent advanced approaches, which are joint models based on graphs, might still face two potential issues:  (i) the uncertainty introduced by constructing graphs based on preliminary intents and slots, which may transfer intent-slot correlation information to incorrect label node destinations, and (ii) direct incorporation of multiple intent labels for each token w.r.t. token-level intent voting might potentially lead to incorrect slot predictions, thereby hurting the overall performance. To address these two issues, we propose a joint model named MISCA. Our MISCA introduces an intent-slot co-attention mechanism and an underlying layer of label attention mechanism. These mechanisms enable MISCA to effectively capture correlations between intents and slot labels, eliminating the need for graph construction. They also facilitate the transfer of correlation information in both directions: from intents to slots and from slots to intents, through multiple levels of label-specific representations, without relying on token-level intent information.  Experimental results show that MISCA outperforms previous models, achieving new state-of-the-art overall accuracy performances on two benchmark datasets MixATIS and MixSNIPS. This highlights the effectiveness of our attention mechanisms.
\end{abstract}

\input{src/intro}

\input{src/related}

\input{src/method}

\input{src/exp}

\input{src/conclusion}

\section*{Limitations}

It should also be emphasized that our intent-slot co-attention mechanism functions independently of token-level intent information. 
This mechanism generates $|\mathrm{L}^{\text{I}}|$ vectors for multiple intent detection (i.e. multi-label classification). In contrast, the token-level intent decoding strategy bases on token classification, using $n$ vector representations. Recall that $\mathrm{L}^{\text{I}}$ is the intent label set, and $n$ is the number of input word tokens. Therefore, integrating the token-level intent decoding strategy into the intent-slot co-attention mechanism is not feasible.

This work primarily focuses on modeling the interactions between intents and slots using intermediate attentive layers. We do not specifically emphasize leveraging label semantics or the meaning of labels in natural language. However, although MISCA consistently outperforms previous models, its performance may be further enhanced by estimating the semantic similarity between words in an utterance and in the labels.

\bibliography{references}
\bibliographystyle{acl_natbib}

\appendix

\end{document}

%% file: src/intro.tex
\begin{figure*}
    \centering
    \includegraphics[scale=0.535]{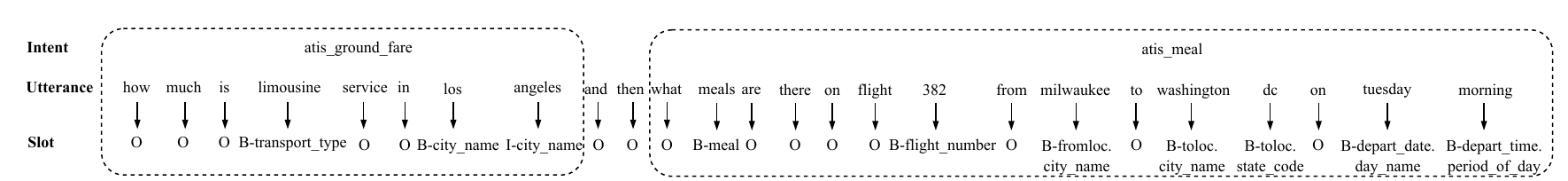}
    \caption{An example of utterance with multiple intents and slots.}
    \label{fig:example}
\end{figure*}

\section{Introduction}

Spoken language understanding (SLU) is a fundamental component in various applications, ranging from virtual assistants to chatbots and intelligent systems. In general, SLU involves two tasks: intent detection to classify the intent of user utterances, and slot filling to extract useful semantic concepts \citep{tur2011spoken}. A common approach to tackling these tasks is through sequence classification for intent detection and sequence labeling for slot filling. Recent research on this topic, recognizing the high correlation between intents and slots, shows that a joint model can improve overall performance by leveraging the inherent dependencies between the two tasks \citep{louvan2020recent, zhang2018joint, weld2022survey}. A number of joint models have been proposed to exploit the correlations between single-intent detection and slot filling tasks, primarily by incorporating attention mechanisms \citep{goo2018slot, li2018self,niu2019novel, qin2019stack, zhang2019joint, chen2019bert,JointIDSF}.

However, in real-world scenarios, users may often express utterances with multiple intents, as illustrated in Figure \ref{fig:example}. This poses a challenge for single-intent systems, potentially resulting in poor performance. Recognizing this challenge, \citet{kim2017two} is the first to explore the detection of multiple intents in a SLU system, followed by \citet{gangadharaiah-narayanaswamy-2019-joint} who first propose a joint framework for multiple intent detection and slot filling. \citet{qin2020agif} and \citet{qin2021gl} have further explored the utilization of graph attention network \citep{velivckovic2017graph} to explicitly model the interaction between predicted intents and slot mentions. Recent state-of-the-art models Co-guiding  \cite{xing2022co} and Rela-Net  \cite{xing2022group} further incorporate the guidance from slot information to aid in intent prediction, and design heterogeneous graphs to facilitate more effective interactions between intents and slots. 

We find that two potential issues might still persist within these existing multi-intent  models: \textbf{(1)} The lack of ``gold'' graphs that accurately capture the underlying relationships and dependencies between intent and slot labels in an utterance. Previous graph-based methods construct a graph for each utterance by predicting preliminary intents and slots  \citep{qin2021gl, xing2022co, xing2022group}. However, utilizing such graphs to update representations of intents and slots might introduce uncertainty as intent-slot correlation information could be transferred to incorrect label node destinations. \textbf{(2)} The direct incorporation of multiple intent labels for each word token to facilitate token-level intent  voting \citep{qin2021gl, xing2022co, xing2022group}. To be more specific, due to the absence of token-level gold intent labels, the models are trained to predict multiple intent labels for each word token in the utterance. 
Utterance-level intents depend on these token-level intents and require at least half of all tokens to support an intent. In Figure \ref{fig:example}, for instance, a minimum of 12 tokens is needed to support the ``atis\_ground\_fare'' intent, while only 8 tokens (enclosed by the left rectangle) support it. Note that each token within the utterance context is associated with a specific intent. Thus, incorporating irrelevant intent representations from each token might potentially lead to incorrect slot predictions, thereby hurting the overall accuracy.

To overcome the above issues, in this paper, we propose a new joint \textbf{m}odel with an \textbf{i}ntent-\textbf{s}lot \textbf{c}o-\textbf{a}ttention mechanism, which we name MISCA, for multi-intent detection and slot filling. 
Equivalently, our novel co-attention mechanism serves as an effective replacement for the graph-based interaction module employed in previous works, eliminating the need for explicit graph construction. By enabling seamless intent-to-slot and slot-to-intent information transfer, our co-attention mechanism facilitates the exchange of relevant information between intents and slots. This novel mechanism not only simplifies the model architecture, but also maintains the crucial interactions between intent and slot representations, thereby enhancing the overall performance.

In addition, MISCA also presents a label attention mechanism as an underlying layer for the co-attention mechanism. This label attention mechanism operates independently of token-level intent information and is designed specifically to enhance the extraction of slot label- and intent label-specific representations. By capturing the characteristics of each intent/slot label, the label attention mechanism helps MISCA obtain a deep understanding and fine-grained information about the semantic nuances associated with different intent and slot labels. This, in turn, ultimately helps improve the overall results of intent detection and slot filling.

Our contributions are summarized as follows: \textbf{(I)} We introduce a novel joint model called MISCA for multiple intent detection and slot filling tasks, which incorporates label attention and intent-slot co-attention mechanisms.\footnote{Our MISCA implementation is publicly available at: \url{https://github.com/VinAIResearch/MISCA}.}
\textbf{(II)}  MISCA effectively captures correlations between intents and slot labels and facilitates the transfer of correlation information in both the intent-to-slot and slot-to-intent directions through multiple levels of label-specific representations.
\textbf{(III)} Experimental results show that our MISCA outperforms previous strong baselines, achieving new state-of-the-art overall accuracies on two benchmark datasets.

\begin{figure*}[!t]
    \centering
    \includegraphics[width=0.975\linewidth]{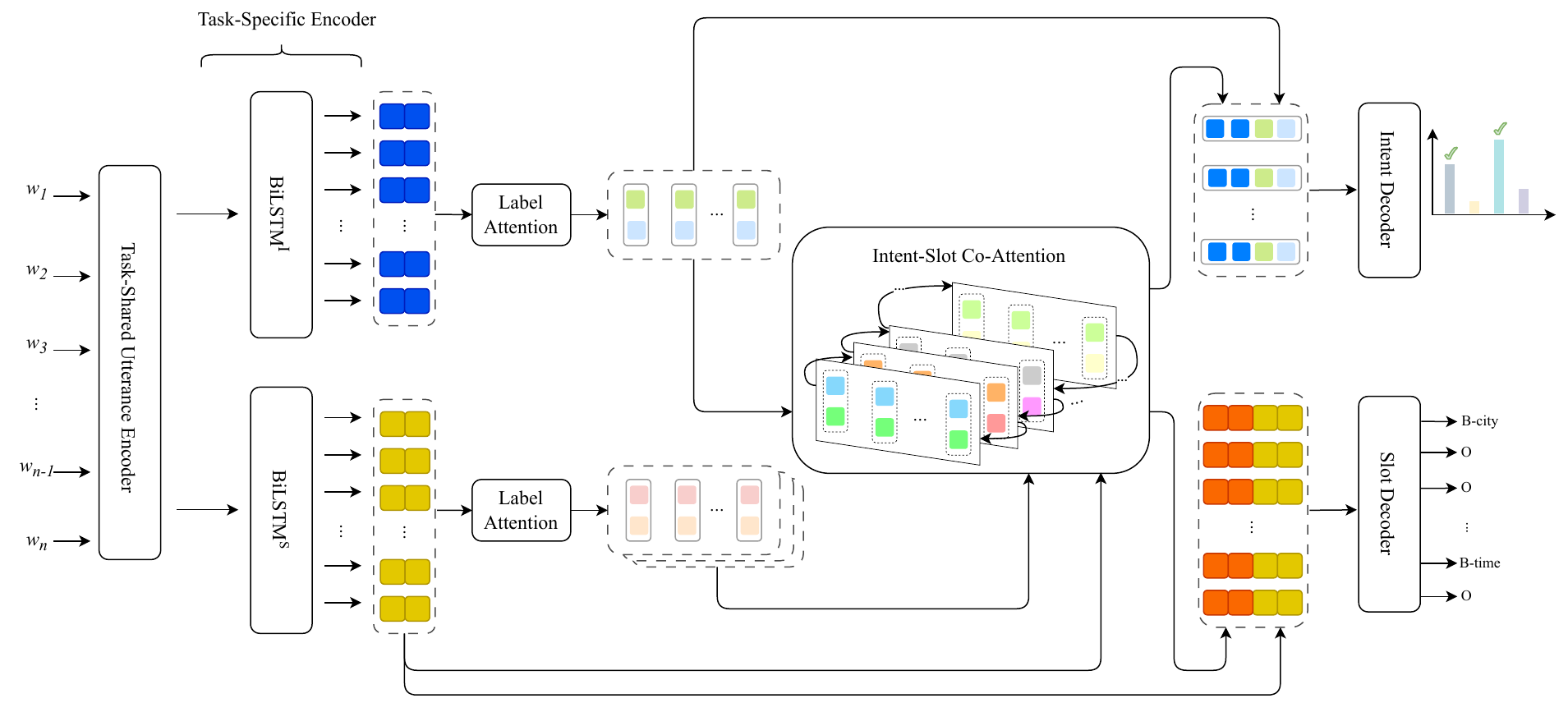}
    \caption{Illustration of the architecture of our joint model MISCA.}
    \label{fig:model}
\end{figure*}

%% file: src/related.tex
\section{Problem Definition and Related Work}

Given an input utterance consisting of $n$ word tokens $w_1, w_2,..., w_n$, the \textit{multiple intent detection} task is a \underline{multi-label classification} problem that predicts multiple intents of the input utterance. Meanwhile, the \textit{slot filling} task can be viewed as a \underline{sequence labeling} problem that predicts a slot label for each token of the input utterance.

\citet{kim2017two} show the significance of the  multiple intents setting in SLU. \citet{gangadharaiah-narayanaswamy-2019-joint} then introduce a joint approach for multiple intent detection and slot filling, which models relationships between slots and intents via a slot-gated mechanism. 
However, this slot-gated mechanism represents multiple intents using only one feature vector, and thus incorporating this feature vector to guide slot filling could lead to incorrect slot predictions.

To generate fine-grained intents information for slot label prediction, \citet{qin2020agif} introduce an adaptive interaction framework based on graph attention networks. However, the autoregressive nature of the framework restricts its ability to use bidirectional information for slot filling. 
To overcome this limitation, \citet{qin2021gl} proposes a global-locally graph interaction network that incorporates both a global graph to model interactions between intents and slots, and a local graph to capture relationships among slots.

More recently, \citet{xing2022co} propose two heterogeneous graphs, namely slot-to-intent and intent-to-slot, to establish mutual guidance between the two tasks based on preliminarily predicted intent and slot labels. Meanwhile, \citet{xing2022group} propose a heterogeneous label graph that incorporates statistical dependencies and hierarchies among labels to generate label embeddings. They leverage both the label embeddings and the hidden states from a label-aware inter-dependent decoding mechanism to construct decoding processes between the two tasks. See a discussion on potential issues of these models in the Introduction. Please find more related work in Section \ref{subsec:baseline}. 

%% file: src/method.tex
\section{Our MISCA model}
\label{sec:method}

Figure \ref{fig:model} illustrates the architecture of our MISCA, which consists of four main components: (i) Task-shared and task-specific utterance encoders, (ii) Label attention, (iii) Intent-slot co-attention, and (iv) Intent and slot decoders.

The encoders component aims to generate intent-aware and slot-aware task-specific feature vectors for intent detection and slot filling, respectively. The label attention component takes these task-specific vectors as input and outputs label-specific feature vectors.
The intent-slot co-attention component utilizes the label-specific vectors and the slot-aware task-specific vectors to simultaneously learn correlations between intent detection and slot filling through multiple intermediate layers. The output vectors generated by this co-attention component are used to construct input vectors for the intent and slot decoders which predict multiple intents and slot labels, respectively.

\subsection{Utterance encoders}
\label{subsubsec:encoder}

Following previous work \cite{qin2020agif, qin2021gl,song2022enhancing}, we employ a task-shared encoder and a task-specific encoder. 

\paragraph{Task-shared encoder:} Given an input utterance consisting of $n$ word tokens $w_1, w_2, ..., w_n$, our task-shared encoder creates a vector $\mathbf{e}_{i}$ to represent the $i^{\text{th}}$ word token $w_i$ by  concatenating contextual word embeddings $\mathbf{e}_{i}^{\text{BiLSTM\textsubscript{word}}}$ and $\mathbf{e}_{i}^{\text{SA}}$, and character-level word embedding $\mathbf{e}_{w_i}^{\text{BiLSTM\textsubscript{char.}}}$:
\begin{eqnarray}
\mathbf{e}_{i} &=& \mathbf{e}_{i}^{\text{BiLSTM\textsubscript{word}}} \oplus \mathbf{e}_{i}^{\text{SA}} \oplus \mathbf{e}_{w_i}^{\text{BiLSTM\textsubscript{char.}}} \label{equation:1} 
\end{eqnarray}

Here, we feed a sequence  $\mathbf{e}_{w_1:w_n}$ of real-valued word embeddings $\mathbf{e}_{w_1}$, $\mathbf{e}_{w_2}$,... $\mathbf{e}_{w_n}$ into a single bi-directional LSTM (BiLSTM\textsubscript{word}) layer \citep{hochreiter1997long} and a single self-attention layer \citep{vaswani2017attention} to produce the contextual feature vectors $\mathbf{e}_{i}^{\text{BiLSTM\textsubscript{word}}}$ and $\mathbf{e}_{i}^{\text{SA}}$, respectively. In addition, the character-level word embedding $\mathbf{e}_{w_i}^{\text{BiLSTM\textsubscript{char.}}}$ is derived by applying another single BiLSTM (BiLSTM\textsubscript{char.}) to the sequence of real-valued embedding representations of characters in each word ${w_i}$, as done in \citet{lample-etal-2016-neural}.

\paragraph{Task-specific encoder:} Our task-specific encoder passes the sequence of vectors $\mathbf{e}_{1:n}$ as input to two different single BiLSTM layers to produce task-specific latent  vectors $\mathbf{e}_{i}^{\text{I}} = \mathrm{BiLSTM}\textsuperscript{I}(\mathbf{e}_{1:n}, i)$ and $\mathbf{e}_{i}^{\text{S}} = \mathrm{BiLSTM}\textsuperscript{S}(\mathbf{e}_{1:n}, i) \in \mathbb{R}^{d_e}$ for intent detection and slot filling, respectively. These task-specific vectors are concatenated to formulate task-specific matrices $\mathbf{E}^{\text{I}}$ and $\mathbf{E}^{\text{S}}$ as follows:

\begin{eqnarray}
\mathbf{E}^{\text{I}} &=& [\mathbf{e}_{1}^{\text{I}}, \mathbf{e}_{2}^{\text{I}},... \mathbf{e}_{n}^{\text{I}}] \in \mathbb{R}^{d_e\times n} \label{equ:E_I}\\
\mathbf{E}^{\text{S}} &=& [\mathbf{e}_{1}^{\text{S}}, \mathbf{e}_{2}^{\text{S}},... \mathbf{e}_{n}^{\text{S}}] \in \mathbb{R}^{d_e\times n} \label{equ:E_S}
\end{eqnarray}

\subsection{Label  attention}
\label{subsec:label}

The word tokens in the input utterance might make different contributions to each of the intent and slot labels \citep{xiao2019label,song2022enhancing}, motivating our extraction of label-specific vectors representing intent and slot labels. In addition, most previous works show that the slot labels might share some semantics of  hierarchical relationships \cite{weld2022survey}, e.g. ``fine-grained'' labels \texttt{toloc.city\_name},  \texttt{toloc.state\_name} and \texttt{toloc.country\_name} can be grouped into a more ``coarse-grained'' label type \texttt{toloc}. 
We thus introduce a hierarchical label attention mechanism, adapting the attention mechanism from \citet{ijcai2020-461-vu}, to take such slot label hierarchy information into extracting the label-specific vectors. 

Formally, our label attention mechanism takes the task-specific matrix (here, $\mathbf{E}^{\text{I}}$ from Equation \ref{equ:E_I} and $\mathbf{E}^{\text{S}}$ from Equation \ref{equ:E_S}) as input and computes a label-specific attention weight matrix (here, $\mathbf{A}^{\text{I}} \in \mathbb{R}^{|\mathrm{L}^{\text{I}}| \times n}$ and $\mathbf{A}^{\text{S},k} \in \mathbb{R}^{|\mathrm{L}^{\text{S},k}| \times n}$  at the $k^{\text{th}}$ hierarchy level of slot labels) as follows:

\begin{eqnarray}
\mathbf{A}^{\text{I}} &=& \mathrm{softmax}\big(\mathbf{B}^{\text{I}} \times  \mathrm{tanh}(\mathbf{D}^{\text{I}} \times \mathbf{E}^{\text{I}})\big) \\
\mathbf{A}^{\text{S},k} &=& \mathrm{softmax}\big(\mathbf{B}^{\text{S},k} \times  \mathrm{tanh}(\mathbf{D}^{\text{S},k} \times \mathbf{E}^{\text{S}})\big)
\end{eqnarray}

\noindent where $\mathrm{softmax}$ is performed at the row level to make sure that the summation of weights in each row is equal to 1;  and $\mathbf{B}^{\text{I}} \in \mathbb{R}^{|\mathrm{L}^{\text{I}}| \times d_a}$, $\mathbf{D}^{\text{I}} \in \mathbb{R}^{d_a\times d_e}$, $\mathbf{B}^{\text{S},k} \in \mathbb{R}^{|\mathrm{L}^{\text{S},k}| \times d_a}$ and $\mathbf{D}^{\text{S},k} \in \mathbb{R}^{d_a\times d_e}$, in which $\mathrm{L}^{\text{I}}$ and $\mathrm{L}^{\text{S},k}$ are the intent label set and the  set of slot  label types at the $k^{\text{th}}$ hierarchy level, respectively. Here, $k \in \{1, 2,..., \ell\}$ where $\ell$ is the number of hierarchy levels of slot labels, and thus $\mathrm{L}^{\text{S},\ell}$ is the set of ``fine-grained'' slot label types (i.e. all original slot labels in the training data). 

After that, label-specific representation matrices $\mathbf{V}^{\text{I}}$ and $\mathbf{V}^{\text{S},k}$ are computed by multiplying the task-specific matrices $\mathbf{E}^{\text{I}}$ and $\mathbf{E}^{\text{S}}$ with the attention weight matrices $\mathbf{A}^{\text{I}}$ and $\mathbf{A}^{\text{S},k}$, respectively, as: 

\begin{eqnarray}
\mathbf{V}^{\text{I}} &=& \mathbf{E}^{\text{I}} \times \big(\mathbf{A}^{\text{I}}\big)^{\top} \label{equation:6}\\ 
\mathbf{V}^{\text{S},k} &=& \mathbf{E}^{\text{S}} \times \big(\mathbf{A}^{\text{S},k}\big)^{\top} \label{equation:7}
\end{eqnarray}

Here, the $j^{\text{th}}$ columns $\mathbf{v}_{j}^{\text{I}}$ from $\mathbf{V}^{\text{I}} \in \mathbb{R}^{d_e \times |\mathrm{L}^{\text{I}}|}$ and $\mathbf{v}_{j}^{\text{S},k}$ from $\mathbf{V}^{\text{S},k} \in \mathbb{R}^{d_e \times |\mathrm{L}^{\text{S},k}|}$ are referred to as vector representations of the input utterance w.r.t. the $j^{\text{th}}$ label in $\mathrm{L}^{\text{I}}$ and $\mathrm{L}^{\text{S},k}$, respectively. 

To capture slot label hierarchy information, at $k \geq 2$, taking $\mathbf{v}_{j}^{\text{S},k-1}$, we compute the probability $\mathrm{p}_{j}^{\text{S},k-1}$ of the  $j^{\text{th}}$ slot label  at the $(k-1)^{\text{th}}$ hierarchy level given the utterance, using a corresponding weight vector $\mathbf{w}_j^{\text{S},k-1} \in \mathbb{R}^{d_e}$ and the $\mathrm{sigmoid}$ function. We  project the vector $\mathbf{p}^{\text{S},k-1}$ of label probabilities $\mathrm{p}_{j}^{\text{S},k-1}$ using a projection matrix $\mathbf{Z}^{\text{S},k-1} \in \mathbb{R}^{d_p\times|\mathrm{L}^{\text{S},k-1}|}$, and then concatenate the projected vector output with each slot label-specific vector of the $k^{\text{th}}$ hierarchy level:

\begin{eqnarray}
\mathrm{p}_{j}^{\text{S},k-1} &=& \mathrm{sigmoid}\big(\mathbf{w}_j^{\text{S},k-1} \cdot \mathbf{v}_{j}^{\text{S},k-1}\big ) \label{equation:8} \\
\mathbf{p}^{\text{S},k-1} &=& [\mathrm{p}_{1}^{\text{S},k-1}, \mathrm{p}_{2}^{\text{S},k-1},..., \mathrm{p}_{|\mathrm{L}^{\text{S},k-1}|}^{\text{S},k-1}]^\top \label{equation:9} \\
\mathbf{v}_{j}^{\text{S},k} &\leftarrow& \mathbf{v}_{j}^{\text{S},k} \oplus  \mathbf{Z}^{\text{S},k-1} \times \mathbf{p}^{\text{S},k-1} \label{equation:10} \\
\mathbf{V}^{\text{S},k} &=& [\mathbf{v}_{1}^{\text{S},k}, \mathbf{v}_{2}^{\text{S},k},..., \mathbf{v}_{|\mathrm{L}^{\text{S},k}|}^{\text{S},k}] \label{equation:11}
\end{eqnarray}

The slot label-specific matrix  $\mathbf{V}^{\text{S},k}$ at $k \geq 2$ is now updated with more ``coarse-grained'' label information from the $(k-1)^{\text{th}}$ hierarchy level.

\subsection{Intent-slot co-attention}
\label{subsec:hier}

Given that intents and slots presented in the same utterance share correlation information \cite{louvan2020recent,weld2022survey}, it is intuitive to consider modeling interactions between them. For instance, utilizing intent context vectors could enhance slot filling, while slot context vectors could improve intent prediction. We thus introduce a novel intent-slot co-attention mechanism that extends the parallel co-attention from \citet{lu2016hierarchical}. Our mechanism 
allows for simultaneous attention to intents and slots through multiple intermediate layers.

Our co-attention mechanism creates a matrix $\mathbf{S}\in \mathbb{R}^{d_s \times n}$ whose each column represents a ``soft'' slot label embedding for each input word token, based on its task-specific feature vector:

\begin{eqnarray}
    \mathbf{S} &=& \mathbf{W}^{\text{S}}\mathrm{softmax}\big(\mathbf{U}^{\text{S}}\mathbf{E}^\text{S}\big)  \label{equation:12}
\end{eqnarray}

\noindent where $\mathbf{W}^{\text{S}} \in \mathbb{R}^{d_s \times (2|\mathrm{L}^{\text{S},\ell}|+1)}$,$\mathbf{U}^{\text{S}} \in \mathbb{R}^{(2|\mathrm{L}^{\text{S},\ell}|+1)\times d_e}$ and $2|\mathrm{L}^{\text{S},\ell}|+1$ is the number of BIO-based slot tag labels (including the ``O'' label) as we formulate the slot filling task as a BIO-based sequence labeling problem.  Recall that $\mathrm{L}^{\text{S},\ell}$ is the set of ``fine-grained'' slot label types without ``B-'' and ``I-'' prefixes, not including the ``O'' label. Here, $\mathrm{softmax}$ is performed at the column level.

Our mechanism takes a sequence of $\ell + 2$   input feature matrices $\mathbf{V}^\text{I}$, $\mathbf{V}^{\text{S},1}$, $\mathbf{V}^{\text{S},2}$,..., $\mathbf{V}^{\text{S},\ell}$, $\mathbf{S}$ (computed as in Equations \ref{equation:6}, \ref{equation:7}, \ref{equation:11}, \ref{equation:12}) to perform intent-slot co-attention. 

\underline{For notation simplification}, the input feature  matrices of our mechanism are orderly referred to as $\mathbf{Q}_1, \mathbf{Q}_2, ..., \mathbf{Q}_{\ell + 2}$, where $\mathbf{Q}_1 = \mathbf{V}^\text{I}$, $\mathbf{Q}_2 = \mathbf{V}^{\text{S},1}$,..., $\mathbf{Q}_{\ell + 1} = \mathbf{V}^{\text{S},\ell}$ and $\mathbf{Q}_{\ell + 2} = \mathbf{S}$; and  $d_t\times m_t$ is the size of the corresponding matrix $\mathbf{Q}_t$ whose each column is referred to as a label-specific vector: 
$d_1 = d_e\ , m_1 = |\mathrm{L}^{\text{I}}|$; $d_2 = d_e\ , m_2 = |\mathrm{L}^{\text{S},1}|$; $d_3 = d_e + d_p\ , m_3 = |\mathrm{L}^{\text{S},2}|$; ...; $d_{\ell + 1}= d_e + d_p\ , m_{\ell + 1} = |\mathrm{L}^{\text{S},\ell}|$; $d_{\ell + 2}=d_s\ , m_{\ell + 2} = n$.

As each intermediate layer's matrix $\mathbf{Q}_{t}$ has different interactions with the previous layer's matrix  $\mathbf{Q}_{t-1}$ and the next  layer's matrix $\mathbf{Q}_{t+1}$, we project $\mathbf{Q}_{t}$  into two vector spaces  to ensure that all label-specific column vectors have the same dimension:

\begin{equation}
    \overrightarrow{\mathbf{Q}}_t = \overrightarrow{\mathbf{W}}_t \mathbf{Q}_t \quad ; \quad
    \overleftarrow{\mathbf{Q}}_t = \overleftarrow{\mathbf{W}}_t \mathbf{Q}_t
\end{equation}

\noindent where $\overrightarrow{\mathbf{W}}_t$ and $\overleftarrow{\mathbf{W}}_t \in \mathbb{R}^{d \times d_t}$ are projection weight matrices;  and thus $\overrightarrow{\mathbf{Q}}_t$ and $\overleftarrow{\mathbf{Q}}_t \in \mathbb{R}^{d \times m_t}$.

We also compute a bilinear attention between two matrices $\mathbf{Q}_{t-1}$ and $\mathbf{Q}_{t}$ to measure the correlation between their corresponding label types: 

\begin{equation}
    \mathbf{C}_t = \mathbf{Q}_{t-1}^\top \mathbf{X}_t \mathbf{Q}_t  \label{equation:15}
\end{equation}

\noindent where $\mathbf{X}_t \in \mathbb{R}^{d_{t-1}\times d_t}$, and thus  $\mathbf{C}_t \in \mathbb{R}^{m_{t-1}\times m_t}$.

Our co-attention mechanism allows the intent-to-slot and slot-to-intent information transfer by computing attentive label-specific representation matrices as follows:

\begin{eqnarray}
    \overleftarrow{\mathbf{H}}_{t} &=& 
    \begin{dcases}
     \mathrm{tanh}(\overleftarrow{\mathbf{Q}}_{t+1}\mathbf{C}_{t+1}^\top + \overleftarrow{\mathbf{Q}}_{t})\ ,\ \text{if } t = \ell +1\\
     \mathrm{tanh}(\overleftarrow{\mathbf{H}}_{t+1}\mathbf{C}_{t+1}^\top + \overleftarrow{\mathbf{Q}}_{t})\ ,\ \text{otherwise}
    \end{dcases} \label{equation:15} \\
    \overrightarrow{\mathbf{H}}_t &=&
    \begin{dcases}
     \mathrm{tanh}(\overrightarrow{\mathbf{Q}}_{t-1}\mathbf{C}_{t} + \overrightarrow{\mathbf{Q}}_t)\ ,\ \text{if } t = 2\\
     \mathrm{tanh}(\overrightarrow{\mathbf{H}}_{t-1}\mathbf{C}_{t} + \overrightarrow{\mathbf{Q}}_t)\ ,\ \text{otherwise}
    \end{dcases} \label{equation:16}
\end{eqnarray}

We use $\overleftarrow{\mathbf{H}}_1 \in \mathbb{R}^{d\times |\mathrm{L}^{\text{I}}|}$ and $\overrightarrow{\mathbf{H}}_{\ell +2} \in \mathbb{R}^{d\times n}$  as computed following Equations \ref{equation:15} and \ref{equation:16} as the  matrix outputs representing intents and slot mentions, respectively. 

\subsection{Decoders}
\label{subsubsec:decoder}

\paragraph{Multiple intent decoder:} We formulate the multiple intent detection task as a multi-label classification problem. We  concatenate $\mathbf{V}^{\text{I}}$ (computed as in Equation \ref{equation:6}) and $\overleftarrow{\mathbf{H}}_1$ (computed following Equation \ref{equation:15}) to create an intent label-specific matrix $\mathbf{H}^{\text{I}} \in \mathbb{R}^{(d_e+d)\times |\mathrm{L}^{\text{I}}|}$ where its $j^{\text{th}}$ column vector $\boldsymbol{v}_j^{\text{I}} \in \mathbb{R}^{d_e+d}$ is referred to as the final vector representation of the input utterance w.r.t. the $j^{\text{th}}$ intent label in $\mathrm{L}^{\text{I}}$. Taking $\boldsymbol{v}_j^{\text{I}}$, we compute the probability $\mathrm{p}_{j}^{\text{I}}$ of the  $j^{\text{th}}$ intent label given the  utterance by using a corresponding weight vector and the $\mathrm{sigmoid}$ function, following Equation \ref{equation:8}. 

We also follow previous works to incorporate an  auxiliary task of predicting the number of intents given the input utterance \cite{chen2022transformer, cheng2022scope, zhu2023dynamic}. In particular, we compute the number $y^{\text{INP}}$ of intents for the input utterance as: $y^{\text{INP}} = \argmax\big(\mathrm{softmax}\big(\mathbf{W}^{\text{INP}}(\mathbf{V}^{\text{I}})^\top\mathbf{w}^{\text{INP}}\big)\big)$, where $\mathbf{W}^{\text{INP}} \in \mathbb{R}^{z\times|\mathrm{L}^{\text{I}}|}$ and $\mathbf{w}^{\text{INP}}\in \mathbb{R}^{d_e}$ are weight matrix and vector, respectively, and $z$ is the maximum number of gold intents for an utterance in the training data. 
We then select the top $y^{\text{INP}}$ highest probabilities $\mathrm{p}_{j}^{\text{I}}$ and consider their corresponding intent labels as the final intent outputs. 

Our intent detection object loss $\mathcal{L}_{\text{ID}}$ is computed as the sum of the  binary cross entropy loss based on the probabilities $\mathrm{p}_{j}^{\text{I}}$ for multiple intent prediction and the multi-class cross entropy loss for predicting the number $y^{\text{INP}}$ of intents.

\paragraph{Slot decoder:} We formulate the slot filling task as a sequence labeling problem based on the BIO scheme. We  concatenate $\mathbf{E}^\text{S}$ (computed as in Equation \ref{equ:E_S}) and $\overrightarrow{\mathbf{H}}_{\ell +2}$  (computed following Equation \ref{equation:16}) to create a slot filling-specific matrix  $\mathbf{H}^{\text{S}} \in \mathbb{R}^{(d_e+d)\times n}$ where its $i^{\text{th}}$ column vector $\boldsymbol{v}_i^{\text{S}} \in \mathbb{R}^{d_e+d}$ is referred to as the final vector representation of the $i^{\text{th}}$ input word w.r.t. slot filling.  We project each $\boldsymbol{v}_i^{\text{S}}$ into the $\mathbb{R}^{2|\mathrm{L}^{\text{S},\ell}|+1}$ vector space by using a project matrix $\mathbf{X}^{\text{S}} \in \mathbb{R}^{(2|\mathrm{L}^{\text{S},\ell}|+1)\times(d_e+d)}$ to obtain output vector $\mathbf{h}_i^{\text{S}} = \mathbf{X}^{\text{S}}\boldsymbol{v}_i^{\text{S}}$. We then feed the output vectors $\mathbf{h}_i^{\text{S}}$ into a linear-chain CRF predictor \citep{lafferty2001conditional} for slot label prediction. 

A cross-entropy loss
$\mathcal{L}_{\text{SF}}$ is calculated for slot filling during training while
the Viterbi algorithm is used for inference.

\subsection{Joint training}
\label{subsec:training}

The final training objective loss $\mathcal{L}$ of our model MISCA is a weighted sum of the intent detection loss  $\mathcal{L}_{\text{ID}}$  and the slot filling loss $\mathcal{L}_{\text{SF}}$: 

\begin{equation}
    \mathcal{L} = \lambda\mathcal{L}_{\text{ID}} + (1 - \lambda)\mathcal{L}_{\text{SF}}
    \label{equation:slu_loss}
\end{equation}

%% file: src/exp.tex
\section{Experimental setup}

\label{sec:exp}
\subsection{Datasets and evaluation metrics}
\label{subsec:data}

We conduct experiments using the ``clean'' benchmarks: MixATIS\footnote{\url{https://github.com/LooperXX/AGIF/tree/master/data/MixATIS_clean}}  \cite{hemphill-etal-1990-atis, qin2020agif} and MixSNIPS\footnote{\url{https://github.com/LooperXX/AGIF/tree/master/data/MixSNIPS_clean}} \cite{coucke2018snips, qin2020agif}. MixATIS contains 13,162, 756 and 828 utterances for training, validation and test, while MixSNIPS contains 39,776, 2,198, and 2,199 utterances for training, validation and test, respectively. We  employ evaluation metrics, including the intent accuracy for multiple intent detection, the F\textsubscript{1} score for slot filling, and the overall accuracy which represents the percentage of utterances whose both intents and slots are all correctly predicted (reflecting real-world scenarios). \textit{Overall accuracy thus is referred to as the main metric for comparison}. 

\subsection{Implementation details}
\label{subsec:setting}

The $\ell$ value is $1$ for MixSNIPS because its slot labels do not share any semantics of hierarchical relationships. On the other hand, in MixATIS, we construct a hierarchy with $\ell = 2$ levels, which include ``coarse-grained'' and ``fine-grained'' slot labels. The ``coarse-grained'' labels, placed at the first level of the hierarchy, are label type prefixes that are shared by the ``fine-grained'' slot labels at the second level of the hierarchy (illustrated by the example in the first paragraph in Section \ref{subsec:label}).

In the encoders component, we set the dimensionality of the self-attention layer output to 256 for both datasets. For the $\mathrm{BiLSTM}\textsubscript{word}$, the dimensionality of the LSTM hidden states is fixed at 64 for MixATIS and 128 for MixSNIPS. Additionally, in $\mathrm{BiLSTM}\textsubscript{char.}$, the LSTM hidden dimensionality is set to 32, while in $\mathrm{BiLSTM}\textsuperscript{I}$ and $\mathrm{BiLSTM}\textsuperscript{S}$, it is set to 128 for both datasets (i.e., $d_e$ is $128 * 2 = 256$). In the label attention and intent-slot co-attention components, we set the following dimensional hyperparameters: $d_a$ = 256, $d_p$ = 32, $d_s$ = 128, and $d$ = 128.

To optimize $\mathcal{L}$, we utilize the AdamW optimizer \citep{loshchilov2017decoupled} and set its initial learning rate to 1e-3, with a batch size of $32$. Following previous work, we randomly initialize the word embeddings and character embeddings in the encoders component. The size of character vector embeddings is set to 32. We perform a grid search to select the word embedding size $\in \{64, 128\}$ and the loss mixture weight $\lambda \in \{0.1, 0.25, 0.5, 0.75, 0.9\}$.

Following previous works \cite{qin2020agif,qin2021gl,xing2022co,xing2022group}, we also experiment with another setting of employing a pre-trained language model (\textbf{PLM}). Here, we replace our task-shared encoder with the RoBERTa\textsubscript{base} model \cite{liu2019roberta}. That is, $\mathbf{e}_{i}$ from Equation \ref{equation:1} is now computed as $\mathbf{e}_{i} = \mathrm{RoBERTa}\textsubscript{base}({w_{1:n}, i})$. For this setting, we perform a grid search to find the AdamW initial learning rate $\in\{$1e-6, 5e-6, 1e-5$\}$ and the weight $\lambda \in \{0.1, 0.25, 0.5, 0.75, 0.9\}$ .

For both the original (i.e. without PLM) and with-PLM settings, we train for 100 epochs and calculate the overall accuracy on the validation set after each training epoch. We select the model checkpoint that achieves the highest overall accuracy on the validation set and use it for evaluation on the test set.

\begin{table*}[ht]
\centering
\begin{tabular}{l|ccc|ccc}
\hline 
\multicolumn{1}{l|}{\multirow{3}{*}{Model}} & \multicolumn{3}{c|}{MixATIS}              & \multicolumn{3}{c}{MixSNIPS} \\
\cline{2-7}
\multicolumn{1}{l|}{}  & Intent & Slot & Overall  & Intent  & Slot  & Overall  \\
\multicolumn{1}{l|}{}  & (Acc.) & (F1) & (Acc.) & (Acc.) & (F1) & (Acc.)  \\
                        \hline
                        
   AGIF \citep{qin2020agif}    &  74.4  &  86.7  &  40.8  &  95.1  &  94.2  &  74.2  \\
    GL-GIN  \cite{qin2021gl}   &  76.3  &  88.3  &  43.5  &  95.6  &  94.9  &   75.4  \\
    SDJN  \cite{chen2022joint}   &  77.1  &  88.2  &  44.6  &  96.5  &  94.4  &  75.7  \\
    GISCo \cite{song2022enhancing}    &  75.0  &  88.5  &  48.2  &  95.5  &  95.0  &   75.9  \\
    SSRAN \cite{cheng2022scope} & 77.9 & 89.4 & 48.9 & \textbf{98.4} & \textbf{95.8} & \underline{77.5} \\
    Rela-Net  \cite{xing2022group}   &  \underline{78.5}  &  \underline{90.1}  &  \underline{52.2}  &  97.6  &  94.7  &   76.1 \\
    Co-guiding  \cite{xing2022co}   &  \textbf{79.1}  &  89.8  &  51.3  &  \underline{97.7}  &  95.1  &   \underline{77.5}  \\
    \hline
    Our MISCA   &  76.7  &  \textbf{90.5}  &  \textbf{53.0}  &  97.3  &  \underline{95.2}  &  \textbf{77.9} \\
    \hline
\end{tabular}
\caption{Obtained results without PLM.  The best score is in \textbf{bold}, while the second best score is in \underline{underline}.}
\label{table:results}
\end{table*}

\subsection{Baselines}
\label{subsec:baseline}

For the first setting without PLM, we compare our MISCA against the following strong baselines: (1) AGIF \citep{qin2020agif}: an adaptive graph interactive framework that facilitates fine-grained intent information transfer for slot prediction; (2) GL-GIN \cite{qin2021gl}: a non-autoregressive global-local graph interaction network; (3) SDJN \cite{chen2022joint}: a weakly supervised approach that utilizes multiple instance learning to formulate multiple intent detection, along with self-distillation techniques; (4) GISCo \cite{song2022enhancing}: an integration of global intent-slot co-occurrence across the entire corpus; (5) SSRAN \cite{cheng2022scope}: a scope-sensitive model that focuses on the intent scope and utilizes the interaction between the two intent detection and slot filling tasks;
(6) Rela-Net \cite{xing2022group}: a model that exploits label typologies and relations through a heterogeneous label graph to represent statistical dependencies and hierarchies in rich relations; and (7) Co-guiding \cite{xing2022co}: a two-stage graph-based framework that enables the two tasks to guide each other using the predicted labels.

For the second setting with PLM, we compare MISCA with the PLM-enhanced variant of the models AGIF, GL-GIN, SSRAN, Rela-Net and Co-guiding. We also compare MISCA with the following PLM-based models: (1) DGIF \cite{zhu2023dynamic}, which leverages the semantic information of labels; (2) SLIM \cite{cai2022slim}, which introduces an explicit map of slots to the corresponding intent; (3) UGEN \cite{wu2022incorporating}, a Unified Generative framework that formulates the joint task as a question-answering problem; and (4) TFMN \cite{chen2022transformer}, a threshold-free intent detection approach without using a threshold. Here, DGIF, SLIM and TFMN are based on BERT \cite{devlin-etal-2019-bert,chen2019bert}, while UGEN is based on T5 \cite{raffel2020exploring}.

\section{Experimental results}
\label{subsec:results}

\subsection{Main results}

\paragraph{Results without PLM:}
Table \ref{table:results} reports the obtained results without PLM on the test set, clearly showing that in general, our MISCA outperforms the previous strong baselines, achieving the highest overall accuracies on both datasets. 

In general, aligning the correct predictions between intent and slot labels is challenging, resulting in the overall accuracy being much lower than the intent accuracy and the F\textsubscript{1} score for slot filling. Compared to the baselines, MISCA achieves better alignment between the two tasks due to our effective co-attention mechanism, while maintaining competitive intent accuracy and slot filling F1 scores (here, MISCA also achieves the highest and second highest F\textsubscript{1} scores for slot filling on MixATIS and MixSNIPS, respectively).
Compared to the previous model Rela-Net, our MISCA obtains a 0.8\% and 1.8\% absolute improvement in overall accuracy on MixATIS and MixSNIPS, respectively. In addition, MISCA also outperforms the previous model Co-guiding by 1.7\% and 0.4\% in overall accuracy on MixATIS and MixSNIPS, respectively. The consistent improvements on both datasets result in a substantial gain of 1.1+\% in the average overall accuracy across the two datasets, compared to both Rela-Net and Co-guiding.

We find that MISCA produces a higher improvement in overall accuracy on MixATIS compared to MixSNIPS. One possible reason is that MISCA leverages the hierarchical structure of slot labels in MixATIS, which is not present in MixSNIPS. 
For example, semantically similar ``fine-grained'' slot labels, e.g. \texttt{fromloc.city\_name},  \texttt{city\_name} and \texttt{toloc.city\_name},  might cause ambiguity  for some baselines  in predicting the correct slot labels. However, these  ``fine-grained'' labels belong to different ``coarse-grained'' types in the slot label hierarchy. Our model could distinguish these ``fine-grained'' labels at a certain intent-to-slot information transfer layer (from the intent-slot co-attention in Section \ref{subsec:hier}), thus enhancing  the performance.

\begin{table}[!t]
\centering
\setlength{\tabcolsep}{0.25em}
\resizebox{7.75cm}{!}{
\begin{tabular}{l|cc}
\hline 
\cline{2-3}
Model & MixA & MixS \\
                        \hline
                        
   AGIF + RoBERTa\textsubscript{base}    &  50.0  &  80.7  \\
   SLIM  \cite{cai2022slim} + BERT  &  47.6  &  84.0  \\
   UGEN    \cite{wu2022incorporating} + T5 &  55.3  &  78.8  \\
   DGIF  \cite{zhu2023dynamic} + BERT  &  50.7  &  84.3  \\
    TFMN  \cite{chen2022transformer} + BERT  &  50.2  &  84.7  \\
    GL-GIN   + RoBERTa\textsubscript{base}  &  53.6  &  82.6  \\
    SSRAN + RoBERTa\textsubscript{base} & 54.4 & 83.1 \\
    Rela-Net  + RoBERTa\textsubscript{base}   &  \underline{58.4}  &  83.8  \\
    Co-guiding  + RoBERTa\textsubscript{base}   &  57.5  & \underline{85.3} \\
    \hline
    MISCA + RoBERTa\textsubscript{base} & \textbf{59.1} & \textbf{86.2} \\
    \hline
\end{tabular}
}
\caption{The overall accuracy with PLM. ``MixA'' and ``MixS'' denote MixATIS and MixSNIPS, respectively.}
\label{table:results_roberta}
\end{table}

\begin{table*}[!t]
\centering
\resizebox{15.5cm}{!}{
\setlength{\tabcolsep}{0.3em}
\begin{tabular}{l|ccc|ccc}
\hline 
\multicolumn{1}{l|}{\multirow{2}{*}{Models}} & \multicolumn{3}{c|}{MixATIS}              & \multicolumn{3}{c}{MixSNIPS}         \\
\cline{2-7}
\multicolumn{1}{l|}{}  & Intent (Acc.) & Slot (F1) & Overall (Acc.) & Intent (Acc.) & Slot (F1) & Overall (Acc.) \\
    \hline
     MISCA   &  \textbf{76.7}  &  \textbf{90.5}  &  \textbf{53.0}  &  \textbf{97.3}  &  \textbf{95.2}  &  \textbf{77.9} \\
    \hdashline
    \ \ \ \ (i) w/o ``slot''  label attention     &  75.1 ($\downarrow$1.6)  &  89.3 ($\downarrow$1.2)  &  49.5 ($\downarrow$3.5) & 96.3 ($\downarrow$1.0) &  94.4 ($\downarrow$0.8) &  73.5 ($\downarrow$4.4)   \\
     \ \ \ \ (ii) w/o co-attention    &  75.3 ($\downarrow$1.4)  &  86.8 ($\downarrow$3.7)  &  44.0 ($\downarrow$9.0) &  95.5 ($\downarrow$1.8) &  94.2 ($\downarrow$1.0) &   72.7 ($\downarrow$5.2)  \\
    \hline
\end{tabular}
}
\caption{Ablation results.}
\label{table:ablation}
\end{table*}

\begin{figure*}[!ht]
    \centering
    \includegraphics[scale=0.645]{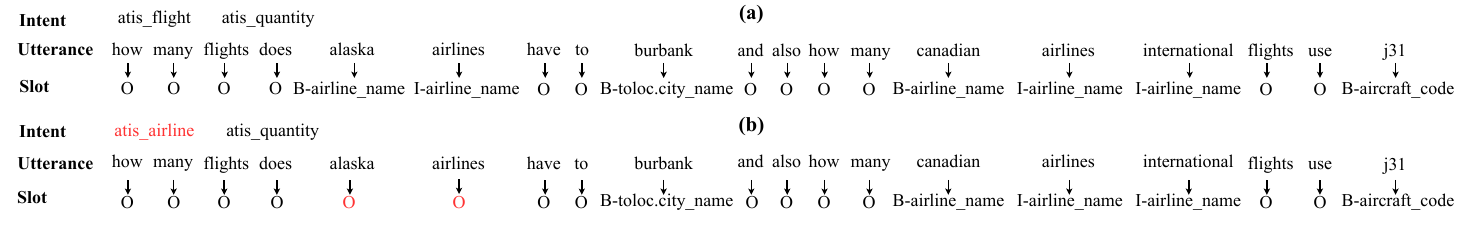}
    \caption{Case study between \textbf{(a)} MISCA and \textbf{(b)} MISCA  w/o co-attention. Red color denotes prediction errors. }
    \label{fig:case_study}
\end{figure*}

\paragraph{State-of-the-art  results with PLM:} Following previous works, we also report the overall accuracy with PLM on the test set. Table \ref{table:results_roberta} presents obtained results comparing our MISCA+RoBERTa with various strong baselines. We find that the PLM notably helps improve the performance of the baselines as well as our MISCA. For example,  RoBERTa helps produce an 6\% accuracy increase  on MixATIS and and an 8\% accuracy increase on MixSNIPS for Rela-Net, Co-guiding and MISCA. Here, MISCA+RoBERTa also consistently outperforms all baselines, producing new state-of-the-art overall accuracies on both datasets: 59.1\% on MixATIS and 86.2\% on MixSNIPS.

\subsection{Ablation study}
\label{subsec:analysis}

We conduct an ablation study with two ablated models: \textbf{(i)} \textbf{w/o ``slot'' label attention} -- This is a variant where we remove all the slot label-specific representation matrices $\mathbf{V}^{\text{S},k}$. That is, our intent-slot co-attention component now only takes 2 input matrices of $\mathbf{Q}_1 = \mathbf{V}^{\text{I}}$ and $\mathbf{Q}_2 = \mathbf{S}$. \textbf{(ii)} \textbf{w/o co-attention} -- This is a  variant where we remove the mechanism component of intent-slot co-attention.  That is, without utilizing $\overleftarrow{\mathbf{H}}_1$ and $\overrightarrow{\mathbf{H}}_{\ell +2}$, we only use  $\mathbf{E}^{\text{S}}$ from the task-specific encoder for the slot decoder, and employ $\mathbf{V}^{\text{I}}$ from the label attention component for the multiple intent decoder (i.e. this can be regarded as a direct adoption of the prior multiple-label decoding approach \cite{ijcai2020-461-vu}). For each ablated model, we also select the model checkpoint that obtains the highest overall accuracy on the validation set to apply to the test.

Table \ref{table:ablation} presents results obtained for both ablated model variants. We find that the model performs substantially poorer when it does not use the slot label-specific  matrices in the intent-slot  co-attention mechanism (i.e. w/o ``slot'' label attention). In this case, the model only considers correlations between intent labels and input word tokens, lacking slot label information necessary to capture intent-slot co-occurrences. We also find that the largest decrease is observed when the intent-slot co-attention mechanism is omitted (i.e. w/o co-attention). Here,   the overall accuracy  drops 9\% on MixATIS and 5.2\% on MixSNIPS.  Both findings strongly indicate the crucial role of the intent-slot  co-attention mechanism in capturing correlations and transferring intent-to-slot and slot-to-intent information between intent and slot labels, leading to notable improvements in the overall accuracy.  

Figure \ref{fig:case_study} showcases a case study to demonstrate the effectiveness of our co-attention mechanism. The baseline MISCA w/o co-attention fails to recognize the slot \texttt{airline\_name} for ``alaska airlines'' and produces an incorrect intent \texttt{atis\_airline}. However, by implementing the intent-slot co-attention mechanism, MISCA accurately predicts both the intent and slot. It leverages information from the slot \texttt{toloc.city\_name} to enhance the probability of the intent \texttt{atis\_flight}, while utilizing intent label-specific vectors to incorporate information about \texttt{airline\_name}. This improvement is achievable due to the effective co-attention mechanism that simultaneously updates intent and slot information without relying on preliminary results from one task to guide the other task.

%% file: src/conclusion.tex
\section{Conclusion}
\label{sec:conclusion}

In this paper, we propose a novel joint model MISCA for multiple intent detection and slot filling tasks. Our MISCA captures correlations between intents and slot labels and transfers the correlation information in both forward and backward directions through multiple levels of label-specific representations.
Experimental results on two benchmark datasets demonstrate the effectiveness of MISCA, which outperforms previous models in both settings: with and without using a pre-trained language model encoder.

%% file: main.bbl
\begin{thebibliography}{38}
\expandafter\ifx\csname natexlab\endcsname\relax\def\natexlab#1{#1}\fi

\bibitem[{Cai et~al.(2022)Cai, Zhou, Mi, and Faltings}]{cai2022slim}
Fengyu Cai, Wanhao Zhou, Fei Mi, and Boi Faltings. 2022.
\newblock {SLIM: Explicit Slot-Intent Mapping with BERT for Joint Multi-Intent
  Detection and Slot Filling}.
\newblock In \emph{Proceedings of ICASSP}, pages 7607--7611.

\bibitem[{Chen et~al.(2022{\natexlab{a}})Chen, Zhou, and Zou}]{chen2022joint}
Lisong Chen, Peilin Zhou, and Yuexian Zou. 2022{\natexlab{a}}.
\newblock Joint multiple intent detection and slot filling via
  self-distillation.
\newblock In \emph{Proceedings of ICASSP}, pages 7612--7616.

\bibitem[{Chen et~al.(2022{\natexlab{b}})Chen, Chen, Zou, Wang, and
  Sun}]{chen2022transformer}
Lisung Chen, Nuo Chen, Yuexian Zou, Yong Wang, and Xinzhong Sun.
  2022{\natexlab{b}}.
\newblock {A Transformer-based Threshold-Free Framework for Multi-Intent NLU}.
\newblock In \emph{Proceedings of COLING}, pages 7187--7192.

\bibitem[{Chen et~al.(2019)Chen, Zhuo, and Wang}]{chen2019bert}
Qian Chen, Zhu Zhuo, and Wen Wang. 2019.
\newblock {BERT for Joint Intent Classification and Slot Filling}.
\newblock \emph{arXiv preprint}, arXiv:1902.10909.

\bibitem[{Cheng et~al.(2022)Cheng, Yang, and Jia}]{cheng2022scope}
Lizhi Cheng, Wenmian Yang, and Weijia Jia. 2022.
\newblock {A Scope Sensitive and Result Attentive Model for Multi-Intent Spoken
  Language Understanding}.
\newblock \emph{arXiv preprint}, arXiv:2211.12220.

\bibitem[{Coucke et~al.(2018)}]{coucke2018snips}
Alice Coucke et~al. 2018.
\newblock {Snips Voice Platform: an embedded Spoken Language Understanding
  system for private-by-design voice interfaces}.
\newblock \emph{arXiv preprint}, arXiv:1805.10190.

\bibitem[{Dao et~al.(2021)Dao, Truong, and Nguyen}]{JointIDSF}
Mai~Hoang Dao, Thinh~Hung Truong, and Dat~Quoc Nguyen. 2021.
\newblock {Intent Detection and Slot Filling for Vietnamese}.
\newblock In \emph{Proceedings of INTERSPEECH}, pages 4698--4702.

\bibitem[{Devlin et~al.(2019)Devlin, Chang, Lee, and
  Toutanova}]{devlin-etal-2019-bert}
Jacob Devlin, Ming-Wei Chang, Kenton Lee, and Kristina Toutanova. 2019.
\newblock {{BERT}: Pre-training of Deep Bidirectional Transformers for Language
  Understanding}.
\newblock In \emph{Proceedings of NAACL}, pages 4171--4186.

\bibitem[{E et~al.(2019)E, Niu, Chen, and Song}]{niu2019novel}
Haihong E, Peiqing Niu, Zhongfu Chen, and Meina Song. 2019.
\newblock {A Novel Bi-directional Interrelated Model for Joint Intent Detection
  and Slot Filling}.
\newblock In \emph{Proceedings of ACL}, pages 5467--5471.

\bibitem[{Gangadharaiah and
  Narayanaswamy(2019)}]{gangadharaiah-narayanaswamy-2019-joint}
Rashmi Gangadharaiah and Balakrishnan Narayanaswamy. 2019.
\newblock {Joint Multiple Intent Detection and Slot Labeling for Goal-Oriented
  Dialog}.
\newblock In \emph{Proceedings of NAACL}, pages 564--569.

\bibitem[{Goo et~al.(2018)Goo, Gao, Hsu, Huo, Chen, Hsu, and
  Chen}]{goo2018slot}
Chih-Wen Goo, Guang Gao, Yun-Kai Hsu, Chih-Li Huo, Tsung-Chieh Chen, Keng-Wei
  Hsu, and Yun-Nung Chen. 2018.
\newblock {Slot-Gated Modeling for Joint Slot Filling and Intent Prediction}.
\newblock In \emph{Proceedings of NAACL}, pages 753--757.

\bibitem[{Hemphill et~al.(1990)Hemphill, Godfrey, and
  Doddington}]{hemphill-etal-1990-atis}
Charles~T. Hemphill, John~J. Godfrey, and George~R. Doddington. 1990.
\newblock {The {ATIS} Spoken Language Systems Pilot Corpus}.
\newblock In \emph{Proceedings of the Third DARPA Speech and Natural Language
  Workshop}, pages 96--101.

\bibitem[{Hochreiter and Schmidhuber(1997)}]{hochreiter1997long}
Sepp Hochreiter and J{\"u}rgen Schmidhuber. 1997.
\newblock {Long Short-Term Memory}.
\newblock \emph{Neural Computation}, 9(8):1735--1780.

\bibitem[{Kim et~al.(2017)Kim, Ryu, and Lee}]{kim2017two}
Byeongchang Kim, Seonghan Ryu, and Gary~Geunbae Lee. 2017.
\newblock Two-stage multi-intent detection for spoken language understanding.
\newblock \emph{Multimedia Tools and Applications}, 76:11377--11390.

\bibitem[{Lafferty et~al.(2001)Lafferty, McCallum, and
  Pereira}]{lafferty2001conditional}
John~D. Lafferty, Andrew McCallum, and Fernando C.~N. Pereira. 2001.
\newblock {Conditional Random Fields: Probabilistic Models for Segmenting and
  Labeling Sequence Data}.
\newblock In \emph{Proceedings of ICML}, page 282–289.

\bibitem[{Lample et~al.(2016)Lample, Ballesteros, Subramanian, Kawakami, and
  Dyer}]{lample-etal-2016-neural}
Guillaume Lample, Miguel Ballesteros, Sandeep Subramanian, Kazuya Kawakami, and
  Chris Dyer. 2016.
\newblock {Neural Architectures for Named Entity Recognition}.
\newblock In \emph{Proceedings of NAACL}, pages 260--270.

\bibitem[{Li et~al.(2018)Li, Li, and Qi}]{li2018self}
Changliang Li, Liang Li, and Ji~Qi. 2018.
\newblock {A Self-Attentive Model with Gate Mechanism for Spoken Language
  Understanding}.
\newblock In \emph{Proceedings of EMNLP}, pages 3824--3833.

\bibitem[{Liu et~al.(2019)Liu, Ott, Goyal, Du, Joshi, Chen, Levy, Lewis,
  Zettlemoyer, and Stoyanov}]{liu2019roberta}
Yinhan Liu, Myle Ott, Naman Goyal, Jingfei Du, Mandar Joshi, Danqi Chen, Omer
  Levy, Mike Lewis, Luke Zettlemoyer, and Veselin Stoyanov. 2019.
\newblock {RoBERTa: A Robustly Optimized BERT Pretraining Approach}.
\newblock \emph{arXiv preprint}, arXiv:1907.11692.

\bibitem[{Loshchilov and Hutter(2019)}]{loshchilov2017decoupled}
Ilya Loshchilov and Frank Hutter. 2019.
\newblock {Decoupled Weight Decay Regularization}.
\newblock In \emph{Proceedings of ICLR}.

\bibitem[{Louvan and Magnini(2020)}]{louvan2020recent}
Samuel Louvan and Bernardo Magnini. 2020.
\newblock {Recent Neural Methods on Slot Filling and Intent Classification for
  Task-Oriented Dialogue Systems: A Survey}.
\newblock In \emph{Proceedings of COLING}, pages 480--496.

\bibitem[{Lu et~al.(2016)Lu, Yang, Batra, and Parikh}]{lu2016hierarchical}
Jiasen Lu, Jianwei Yang, Dhruv Batra, and Devi Parikh. 2016.
\newblock {Hierarchical Question-Image Co-Attention for Visual Question
  Answering}.
\newblock In \emph{Proceedings of NIPS}.

\bibitem[{Qin et~al.(2019)Qin, Che, Li, Wen, and Liu}]{qin2019stack}
Libo Qin, Wanxiang Che, Yangming Li, Haoyang Wen, and Ting Liu. 2019.
\newblock {A Stack-Propagation Framework with Token-Level Intent Detection for
  Spoken Language Understanding}.
\newblock In \emph{Proceedings of EMNLP-IJCNLP}, pages 2078--2087.

\bibitem[{Qin et~al.(2021)Qin, Wei, Xie, Xu, Che, and Liu}]{qin2021gl}
Libo Qin, Fuxuan Wei, Tianbao Xie, Xiao Xu, Wanxiang Che, and Ting Liu. 2021.
\newblock {{GL}-{GIN}: Fast and Accurate Non-Autoregressive Model for Joint
  Multiple Intent Detection and Slot Filling}.
\newblock In \emph{Proceedings of ACL-IJCNLP}, pages 178--188.

\bibitem[{Qin et~al.(2020)Qin, Xu, Che, and Liu}]{qin2020agif}
Libo Qin, Xiao Xu, Wanxiang Che, and Ting Liu. 2020.
\newblock {{AGIF}: An Adaptive Graph-Interactive Framework for Joint Multiple
  Intent Detection and Slot Filling}.
\newblock In \emph{Findings of the ACL: EMNLP}, pages 1807--1816.

\bibitem[{Raffel et~al.(2020)Raffel, Shazeer, Roberts, Lee, Narang, Matena,
  Zhou, Li, and Liu}]{raffel2020exploring}
Colin Raffel, Noam Shazeer, Adam Roberts, Katherine Lee, Sharan Narang, Michael
  Matena, Yanqi Zhou, Wei Li, and Peter~J Liu. 2020.
\newblock {Exploring the Limits of Transfer Learning with a Unified
  Text-to-Text Transformer}.
\newblock \emph{The Journal of Machine Learning Research}, 21(1):5485--5551.

\bibitem[{Song et~al.(2022)Song, Yu, Quangang, Yubin, Liu, and
  Xu}]{song2022enhancing}
Mengxiao Song, Bowen Yu, Li~Quangang, Wang Yubin, Tingwen Liu, and Hongbo Xu.
  2022.
\newblock {Enhancing Joint Multiple Intent Detection and Slot Filling with
  Global Intent-Slot Co-occurrence}.
\newblock In \emph{Proceedings of EMNLP}, pages 7967--7977.

\bibitem[{Tur and De~Mori(2011)}]{tur2011spoken}
Gokhan Tur and Renato De~Mori. 2011.
\newblock \emph{Spoken Language Understanding: Systems for Extracting Semantic
  Information from Speech}.
\newblock John Wiley \& Sons, Inc.

\bibitem[{Vaswani et~al.(2017)Vaswani, Shazeer, Parmar, Uszkoreit, Jones,
  Gomez, Kaiser, and Polosukhin}]{vaswani2017attention}
Ashish Vaswani, Noam Shazeer, Niki Parmar, Jakob Uszkoreit, Llion Jones,
  Aidan~N Gomez, \L~ukasz Kaiser, and Illia Polosukhin. 2017.
\newblock {Attention is All you Need}.
\newblock In \emph{Proceedings of NIPS}.

\bibitem[{Veli{\v{c}}kovi{\'c} et~al.(2018)Veli{\v{c}}kovi{\'c}, Cucurull,
  Casanova, Romero, Lio, and Bengio}]{velivckovic2017graph}
Petar Veli{\v{c}}kovi{\'c}, Guillem Cucurull, Arantxa Casanova, Adriana Romero,
  Pietro Lio, and Yoshua Bengio. 2018.
\newblock {Graph Attention Networks}.
\newblock In \emph{Proceedings of ICLR}.

\bibitem[{Vu et~al.(2020)Vu, Nguyen, and Nguyen}]{ijcai2020-461-vu}
Thanh Vu, Dat~Quoc Nguyen, and Anthony Nguyen. 2020.
\newblock {A Label Attention Model for ICD Coding from Clinical Text}.
\newblock In \emph{Proceedings of IJCAI-20}, pages 3335--3341.

\bibitem[{Weld et~al.(2022)Weld, Huang, Long, Poon, and Han}]{weld2022survey}
Henry Weld, Xiaoqi Huang, Siqu Long, Josiah Poon, and Soyeon~Caren Han. 2022.
\newblock {A Survey of Joint Intent Detection and Slot Filling Models in
  Natural Language Understanding}.
\newblock \emph{ACM Computing Surveys}, 55(8):1--38.

\bibitem[{Wu et~al.(2022)Wu, Wang, Zhang, Chen, and
  Zhang}]{wu2022incorporating}
Yangjun Wu, Han Wang, Dongxiang Zhang, Gang Chen, and Hao Zhang. 2022.
\newblock {Incorporating Instructional Prompts into a Unified Generative
  Framework for Joint Multiple Intent Detection and Slot Filling}.
\newblock In \emph{Proceedings of COLING}, pages 7203--7208.

\bibitem[{Xiao et~al.(2019)Xiao, Huang, Chen, and Jing}]{xiao2019label}
Lin Xiao, Xin Huang, Boli Chen, and Liping Jing. 2019.
\newblock {Label-Specific Document Representation for Multi-Label Text
  Classification}.
\newblock In \emph{Proceedings of EMNLP-IJCNLP}, pages 466--475.

\bibitem[{Xing and Tsang(2022{\natexlab{a}})}]{xing2022co}
Bowen Xing and Ivor Tsang. 2022{\natexlab{a}}.
\newblock {Co-guiding Net: Achieving Mutual Guidances between Multiple Intent
  Detection and Slot Filling via Heterogeneous Semantics-Label Graphs}.
\newblock In \emph{Proceedings of EMNLP}, pages 159--169.

\bibitem[{Xing and Tsang(2022{\natexlab{b}})}]{xing2022group}
Bowen Xing and Ivor Tsang. 2022{\natexlab{b}}.
\newblock {Group is better than individual: Exploiting Label Topologies and
  Label Relations for Joint Multiple Intent Detection and Slot Filling}.
\newblock In \emph{Proceedings of EMNLP}, pages 3964--3975.

\bibitem[{Zhang et~al.(2019{\natexlab{a}})Zhang, Li, Du, Fan, and
  Yu}]{zhang2018joint}
Chenwei Zhang, Yaliang Li, Nan Du, Wei Fan, and Philip Yu. 2019{\natexlab{a}}.
\newblock {Joint Slot Filling and Intent Detection via Capsule Neural
  Networks}.
\newblock In \emph{Proceedings of ACL}, pages 5259--5267.

\bibitem[{Zhang et~al.(2019{\natexlab{b}})Zhang, Zhang, Chen, and
  Zhang}]{zhang2019joint}
Zhichang Zhang, Zhenwen Zhang, Haoyuan Chen, and Zhiman Zhang.
  2019{\natexlab{b}}.
\newblock {A Joint Learning Framework With BERT for Spoken Language
  Understanding}.
\newblock \emph{Ieee Access}, 7:168849--168858.

\bibitem[{Zhu et~al.(2023)Zhu, Xu, Cheng, Song, and Zou}]{zhu2023dynamic}
Zhihong Zhu, Weiyuan Xu, Xuxin Cheng, Tengtao Song, and Yuexian Zou. 2023.
\newblock {A Dynamic Graph Interactive Framework with Label-Semantic Injection
  for Spoken Language Understanding}.
\newblock In \emph{Proceedings of ICASSP}, pages 1--5.

\end{thebibliography}
